\documentclass[10pt,twocolumn,letterpaper]{article}

\usepackage[pagenumbers]{iccv}      %

\definecolor{iccvblue}{rgb}{0.21,0.49,0.74}
\usepackage{graphicx}
\usepackage{amsmath}
\usepackage{amssymb}
\usepackage{tabularx,booktabs}
\newcolumntype{Y}{>{\centering\arraybackslash}X} %

\usepackage{array}
\usepackage{subcaption}

\usepackage[pagebackref,breaklinks,colorlinks]{hyperref}
\usepackage[capitalize]{cleveref}

\usepackage{comment}
\usepackage{amsmath,amssymb} %
\usepackage{wrapfig} %
\usepackage{color}

\usepackage{mathtools}
\usepackage[accsupp]{axessibility}  %

\usepackage[utf8]{inputenc} %
\usepackage[T1]{fontenc}    %
\usepackage{xr-hyper}
\usepackage{url}            %
\usepackage{booktabs}       %
\usepackage{xcolor,colortbl}         %
\usepackage{caption}

\usepackage{multirow}
\usepackage{enumitem}

\usepackage[normalem]{ulem}

\usepackage{subcaption}
\usepackage{float}
\usepackage{lscape}     

\usepackage{xspace}
\usepackage{setspace}
\usepackage{pifont}
\usepackage{comment}

\usepackage{enumitem}%

\title{DreamTexture: Shape from Virtual Texture with Analysis by Augmentation}

\author{
    Ananta R. Bhattarai$^{1}$
    \and
    Xingzhe He$^{2}$
    \and
    Alla Sheffer$^{2}$
    \and
    Helge Rhodin$^{1, 2}$ 
    \and\\
    $^1$Bielefeld University \quad
    $^2$University of British Columbia
}
\begin{document}

\newcommand{\R}{\mathbb{R}}

\newcommand{\va}{\mathbf{a}}
\newcommand{\vb}{\mathbf{b}}
\newcommand{\vc}{\mathbf{c}}
\newcommand{\vd}{\mathbf{d}}
\newcommand{\ve}{\mathbf{e}}
\newcommand{\vf}{\mathbf{f}}
\newcommand{\vg}{\mathbf{g}}
\newcommand{\vh}{\mathbf{h}}
\newcommand{\vi}{\mathbf{i}}
\newcommand{\vj}{\mathbf{j}}
\newcommand{\vk}{\mathbf{k}}
\newcommand{\vl}{\mathbf{l}}
\newcommand{\vm}{\mathbf{m}}
\newcommand{\vn}{\mathbf{n}}
\newcommand{\vo}{\mathbf{o}}
\newcommand{\vp}{\mathbf{p}}
\newcommand{\vq}{\mathbf{q}}
\newcommand{\vr}{\mathbf{r}}
\newcommand{\vt}{\mathbf{t}}
\newcommand{\vu}{\mathbf{u}}
\newcommand{\vv}{\mathbf{v}}
\newcommand{\vw}{\mathbf{w}}
\newcommand{\vx}{\mathbf{x}}
\newcommand{\vy}{\mathbf{y}}
\newcommand{\vz}{\mathbf{z}}

\newcommand{\mA}{\mathbf{A}}
\newcommand{\mB}{\mathbf{B}}
\newcommand{\mC}{\mathbf{C}}
\newcommand{\mD}{\mathbf{D}}
\newcommand{\mE}{\mathbf{E}}
\newcommand{\mF}{\mathbf{F}}
\newcommand{\mG}{\mathbf{G}}
\newcommand{\mH}{\mathbf{H}}
\newcommand{\mI}{\mathbf{I}}
\newcommand{\mJ}{\mathbf{J}}
\newcommand{\mK}{\mathbf{K}}
\newcommand{\mL}{\mathbf{L}}
\newcommand{\mM}{\mathbf{M}}
\newcommand{\mN}{\mathbf{N}}
\newcommand{\mO}{\mathbf{O}}
\newcommand{\mP}{\mathbf{P}}
\newcommand{\mQ}{\mathbf{Q}}
\newcommand{\mR}{\mathbf{R}}
\newcommand{\mS}{\mathbf{S}}
\newcommand{\mT}{\mathbf{T}}
\newcommand{\mU}{\mathbf{U}}
\newcommand{\mV}{\mathbf{V}}
\newcommand{\mW}{\mathbf{W}}
\newcommand{\mX}{\mathbf{X}}
\newcommand{\mY}{\mathbf{Y}}
\newcommand{\mZ}{\mathbf{Z}}

\newcommand{\cA}{\mathcal A}
\newcommand{\cB}{\mathcal B}
\newcommand{\cC}{\mathcal C}
\newcommand{\cD}{\mathcal D}
\newcommand{\cE}{\mathcal E}
\newcommand{\cF}{\mathcal F}
\newcommand{\cG}{\mathcal G}
\newcommand{\cH}{\mathcal H}
\newcommand{\cI}{\mathcal I}
\newcommand{\cJ}{\mathcal J}
\newcommand{\cK}{\mathcal K}
\newcommand{\cL}{\mathcal L}
\newcommand{\cM}{\mathcal M}
\newcommand{\cN}{\mathcal N}
\newcommand{\cO}{\mathcal O}
\newcommand{\cP}{\mathcal P}
\newcommand{\cQ}{\mathcal Q}
\newcommand{\cR}{\mathcal R}
\newcommand{\cS}{\mathcal S}
\newcommand{\cT}{\mathcal T}
\newcommand{\cU}{\mathcal U}
\newcommand{\cV}{\mathcal V}
\newcommand{\cW}{\mathcal W}
\newcommand{\cX}{\mathcal X}
\newcommand{\cY}{\mathcal Y}
\newcommand{\cZ}{\mathcal Z}

\newcommand{\bR}{\mathbb{R}}
\newcommand{\mx}{\mathbf{x}}
\newcommand{\mj}{\mathbf{j}}
\newcommand{\mb}{\mathbf{b}}
\newcommand{\vmu}{\mathbf{\mu}}

\twocolumn[{%
\renewcommand\twocolumn[1][]{#1}%
\maketitle
\newcommand{\vtextheight}{3cm}
\newcommand{\teaserfigw}{0.19}
\newcommand{\teasertextw}{0.17}
\centering
\vspace{-0.4cm}
\includegraphics[width=0.99\textwidth]{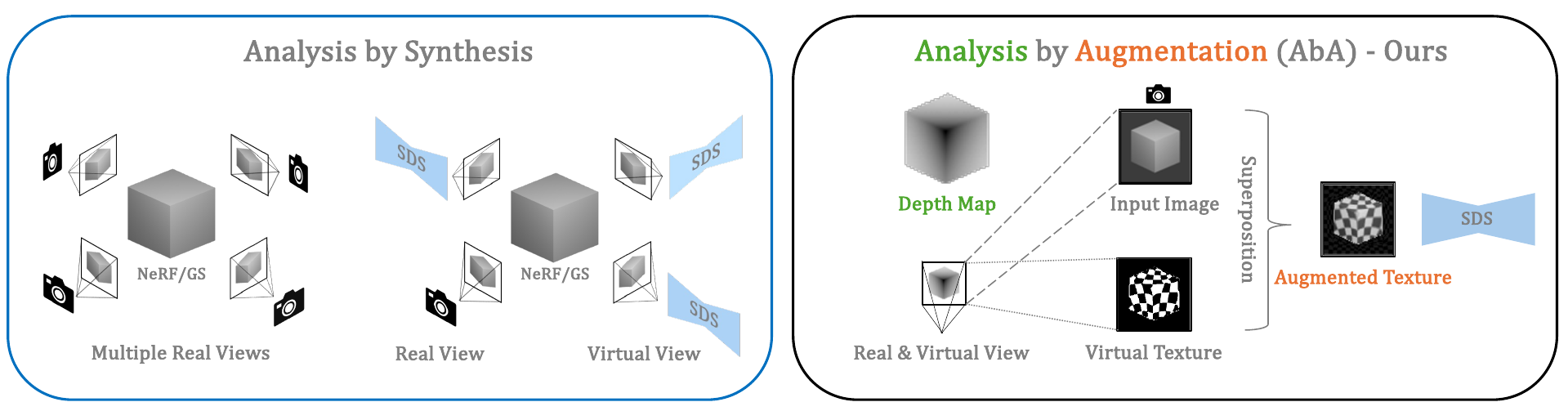}

\captionof{figure}{
    {\bf Teaser.}  While existing analysis-by-synthesis approaches (left) rely on volumetric rendering of multiple real or virtual views,
     \emph{DreamTexture} (right) takes one image as input, augments a virtual texture, and optimizes the output depth map by aligning the real and virtual shape cues with a pre-trained image prior via Score Distillation Sampling (SDS).
    This \emph{analysis by augmentation} (AbA) approach reconstructs depth  without any 3D supervision and unlike classical shape from texture, it applies to both textured and textureless objects.
}
\vspace{1em}
\label{fig:teaser}
}]
\begin{abstract}
DreamFusion established a new paradigm for unsupervised 3D reconstruction from virtual views by combining advances in generative models and differentiable rendering. However, the underlying multi-view rendering, along with supervision from large-scale generative models, is computationally expensive and under-constrained.
We propose DreamTexture, a novel Shape-from-Virtual-Texture approach that leverages monocular depth cues to reconstruct 3D objects. Our method textures an input image by aligning a virtual texture with the real depth cues in the input, exploiting the inherent understanding of monocular geometry encoded in modern diffusion models. We then reconstruct depth from the virtual texture deformation with a new conformal map optimization, which alleviates memory-intensive volumetric representations.
Our experiments reveal that generative models possess an understanding of monocular shape cues, which can be extracted by augmenting and aligning texture cues---a novel monocular reconstruction paradigm that we call Analysis by Augmentation.

\end{abstract}
\section{Introduction}

Unsupervised 3D reconstruction 
is now possible without restrictive assumptions on object class, texture regularity, and shading by
methods such as DreamFusion~\cite{poole2022dreamfusion}. Figure~\ref{fig:teaser} shows how their \emph{virtual \textbf{multi-view} reconstruction} renders virtual views and optimizes the 3D scene parameters to look plausible---such that a pre-trained image prior gives a high likelihood for each view. However, the utilized image priors and the rendering of 3D scenes to multiple views are costly.

To the best of our knowledge, \emph{virtual \textbf{single-view} reconstruction} has not been attempted. The extent to which state-of-the-art generative models understand monocular shape cues and how these can be leveraged remains unexplored. The key benefits of a monocular solution would be alleviating the costly volumetric rendering with NeRFs~\cite{mildenhall2020nerf} or the memory-intensive Gaussian splitting (GS)~\cite{kerbl2023gaussian} and providing additional cues beyond multi-view constraints.

\begin{table}[t!]
    \centering
    \renewcommand{\arraystretch}{1}  %
    \begin{tabular}{cccc}
        \hspace{0.5cm} Input & 
        \hspace{0.4cm} Unaligned & 
        \hspace{0.3cm} Partially & 
        \hspace{0cm} Aligned \\
        \hspace{0.5cm} Image & 
        \hspace{0.4cm} & 
        \hspace{0.3cm} Aligned & 
        \hspace{0cm}  \\
        \multicolumn{4}{c}{\includegraphics[trim={0 0 0 0}, clip, width=0.99\columnwidth]{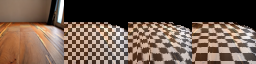}} \\
    \end{tabular}
    \captionof{figure}{\textbf{Analysis by Augmentation.} The virtual texture is augmented and incrementally aligned with the perspective cues in the image (left to right: initialization, 1k iterations, 11k iterations).}
    \label{fig:alignment}
\end{table}

Our new \emph{analysis-by-augmentation} (AbA) principle 
reconstructs from a single view by superimposing a virtual object on the real image.
Likewise, virtual 3D objects must be perfectly registered in the real world for giving humans a realistic augmented reality experience, we leverage a pre-trained image prior to quantify the alignment of our virtual augmentation on the input image.
Our focus is on using texture cues, for which Figure~\ref{fig:alignment} shows the alignment process.

Our \emph{shape-from-virtual-texture} reconstruction (SVT) inverts the notion of classical shape-from-texture (SfT). Traditionally, SfT requires an image of a regularly textured object as input, which is rare in practice~\cite{witkin1981recovering}. By contrast, we generate a virtual object that, when textured with a regular pattern, aligns with the superimposed monocular depth cues in the input image. 
It exploits that the image prior has learned how multiple monocular depth cues appear and align on a particular object.%
\footnote{One could say that the prior has learned the 3D object shape from 2D, but this is rather a philosophical discourse. We only require that the prior recognizes cues that are consistent and rejects inconsistent ones.}
An image prior assigns a lower likelihood to misaligned textures, which we utilize as a loss for 3D reconstruction.
Notably, this also applies to 3D from text by \emph{dreaming up} the input image with a text-conditioned generative model.

A key difficulty in realizing the proposed SVT is generating texture cues that allow gradients to flow from the image prior via the virtual texture to the 3D object shape.
To this end, we rely on conformal maps, which have been extensively studied in the computer graphics literature for texturing a curved object with a regular pattern while minimizing distortion. We re-deploy these principles to find the object shape that best reverts the texture distortion recovered by the image prior.

More precisely, DreamTexture leverages a depth map as a 3D representation and textures it with an associated mesh and hierarchical texture coordinate field. This form has a small memory footprint, and rendering is trivial compared to NeRF or Gaussian Splatting. %
In Stage I, texture coordinates are aligned in 2D using the SDS loss~\cite{poole2022dreamfusion}.
Stage II 
optimizes depth to minimize 2D-to-3D deformation using an angle-preserving loss from conformal maps~\cite{10.1145/566654.566590,sheffer2007mesh}.

Closely related but different in nature is the explicit decomposition of the image into 3D shape, albedo without lighting effects, and a simple shading model with randomized light, as done by DreamFusion~\cite{poole2022dreamfusion} and RealFusion~\cite{MelasKyriazi2023RealFusion3R}. The \emph{analysis-by-synthesis} approach by RealFusion uses a reconstruction objective to minimize the difference between input image and rendering and relies on virtual multi-view constraints because shading cues are insufficient on their own. In contrast, our method superimposes multiple cues on a single view without requiring a perfect decomposition of all factors. We replace decomposition with augmentation and the difference loss with a measure for alignment through the image likelihood.
This enables us to utilize texture cues even if the object to be reconstructed is not regularly textured or provides complex, irreproducible shading cues. Our analysis-by-augmentation approach differs in both construction and effect, enabling generative reconstruction from a single virtual view.

To summarize, our main contributions are:
\begin{itemize}
    \item Establishing the principle of analysis-by-augmentation through the augmentation of monocular texture cues.
    \item Developing a two-stage SVT method by i) tailoring texture coordinate parameterizations to align virtual texture cues using diffusion priors, and ii) reframing least squares conformal mapping (LSCM) for depth optimization.
\end{itemize}

\vspace{0.1cm}
Our qualitative and quantitative evaluation show that our monocular shape-from-virtual-texture approach excels in cases where existing virtual multi-view approaches~\cite{poole2022dreamfusion,MelasKyriazi2023RealFusion3R} struggle due to a lack of image features. %
The AbA principle is scalable without 3D supervision, and we foresee new applications, such as augmenting logos on real images and temporal reconstruction using video diffusion models.

\section{Related Work}

Generative models have introduced new ways to reconstruct 3D objects from virtual views. In the following, we review the different approaches, examine how they resemble multi-view methods in classical computer vision, and outline how monocular shape cues remain underutilized.

\paragraph{Multi-View Image Reconstruction.}
Neural Radiance Fields (NeRFs)~\cite{mildenhall2020nerf, lerf2023, martinbrualla2021nerfw, barron2021mipnerf} have brought dense multi-view reconstruction into the era of neural networks, while Gaussian splatting~\cite{kerbl2023gaussian, huang2024, lyu2024, yang2024gaussianobject} has reinforced the value of classical geometric scene representations. In either case, a differentiable renderer implementing ray tracing or rasterization is used to render each view and backpropagate image reconstruction errors to the scene representation.
While these methods offer high-fidelity reconstruction, they require images taken from dozens of views, and optimizing the high-dimensional scene parameters is a costly, offline process, making them impractical.

Learning-based approaches shift this cost to the training stage~\cite{erkocc2023hyperdiffusion, Shue20223DNF}, optimizing the network to predict a 3D scene representation that explains all the training views, with also some learning the rendering step~\cite{rhodin2018unsupervised, eslami2018neural, rhodin2019neural}. However, much like approaches that utilize ground truth 3D annotation for supervised learning~\cite{Ranftl2020MiDaS, Fu2018DORN, Yao2018MVSNet, Bhat2021AdaBins, Huang2018DeepMVS, Yao2019RMVSNet}, including those using GANs~\cite{wu2016learning, fey2020adversarial, chen2019learning, chan2021piGAN} and diffusion models~\cite{zeng2022lion, lee2023diffusion, wu2024sin3dm, nichol2022pointe, luo2021diffusion, cheng2023sdfusion, li2023generalized}, they struggle to generalize due to the limited availability of multi-view recordings. Even stereo footage is rare, and exploiting symmetry for this purpose in single images is effective only for quasi-symmetric objects such as faces~\cite{wu2020unsupervised}.

\paragraph{Virtual Multi-view Reconstruction.} Generative image models learn what constitutes a realistic image. When evaluated on a synthetic rendering, this provides a loss for reconstruction quality that can replace the real image reconstruction loss in classical multi-view methods. In the extreme case, reconstruction is achieved solely by rendering randomly generated views, as established
by DreamFusion~\cite{poole2022dreamfusion} and later followed by~\cite{lorraine2023att3d, lin2022magic3d, metzer2022latentnerf, wang2023prolificdreamer, liu2023hifa, du2023score}, using SDS on image diffusion models.
This enables monocular reconstruction by pairing one~\cite{MelasKyriazi2023RealFusion3R, deng2023nerdi} or multiple real input images~\cite{yoo2023dreamsparse, raj2023dreambooth3d} with multiple virtual views.
The earliest models using virtual views were GAN-based~\cite{niemeyer2021giraffe, deng2022gram, nguyenphuoc2019hologan, chan2022eg3d, gu2022stylenerf}, where the discriminator takes the role of the image prior and the generator includes an explicit or learned rendering step. 
Irrespective of the backbone used, whether the parameters of a single scene
or a generative model, they all share a common underlying 3D representation and volumetric novel-view renderer, similar to the classical multi-view approaches discussed in the previous paragraph.
Our work alleviates the novel-view rendering by relying solely on single-view monocular cues.

Some recent methods include pre-trained depth estimators~\cite{xu2023neurallift360, tang2023makeit3d} and 3D priors~\cite{qian2024magic123, xu2023dream3d, ding2023text, li2023sweetdreamer}, which counter ambiguities such as the Janus problem but requires massive labeled 3D datasets. By contrast, our method utilizes monocular cues in an unsupervised setting---without leveraging 3D data.

\paragraph{Towards Virtual Single-View Reconstruction.}
To the best of our knowledge, all attempts at 3D reconstruction with an image prior rely on novel view synthesis.
DreamFusion~\cite{poole2022dreamfusion} and RealFusion~\cite{MelasKyriazi2023RealFusion3R} synthesize shading, which provides a virtual monocular cue, but only as auxiliary information within virtual multi-view analysis by synthesis.
In contrast, our virtual SfT method extracts the 3D understanding that the model derives from other monocular cues, such as silhouette and shading, through the new concept of analysis-by-augmentation.
We therefore utilize monocular cues as in classical shape-from-X methods, where X includes shading~\cite{horn1989shape, zhang1999shape}, texture~\cite{witkin1981recovering}, and silhouette~\cite{koenderink1984occluding}.

The closest existing SfT method~\cite{verbin2020toward} leverages generative models to infer 3D shape from texture cues in real images by learning a texture generator and discriminator for each input image. Thus, like classical shape-from-texture~\cite{lobay2006shape, malik1997computing}, it applies only to input images that are regularly textured. In contrast, our texture cues are virtual and can reconstruct untextured or irregularly textured objects.

\section{Preliminaries and Notation}
\label{sec:preliminary}

This section introduces the core methods we build upon. 

\paragraph{Score Distillation Sampling.}
Given a 2D image $\mX$, rendered from a differentiable representation with parameters $\theta$, score distillation sampling (SDS)~\cite{poole2022dreamfusion} utilizes a pre-trained diffusion model $\phi$ to optimize $\theta$ via gradient descent. Specifically, a gradient towards a more likely image is found from the noise $\epsilon_{\phi}$ predicted by $\phi$ given a noisy image $\mX_t$, text embedding $c$, and the noise level $t$,
\begin{equation}
    \nabla_{\theta} \mathcal{L}_{\text{SDS}}(\mathbf{X}, c) = \mathop{\mathbb{E}}_{\epsilon, t} \left[ w(t) \left( \epsilon_{\phi}(\mathbf{X}_t; c, t) - \epsilon \right) \frac{\partial \mathbf{X}}{\partial \theta} \right],
\end{equation}
where $w(t)$ is a weighting function and $\epsilon \sim \mathcal{N}(\mathbf{0}, \mathbf{I})$.
In DreamFusion~\cite{poole2022dreamfusion}, the SDS loss matches the 2D renderings from random angles with the text prompt. In our work, we optimize texture coordinates using SDS to align an augmented virtual texture with shape cues in the input image.

\paragraph{Least Squares Conformal Mapping.} 
Given a 3D mesh $\mathcal{T}$ with each triangle $T \in \mathcal{T}$ having vertex locations $\left((x_1,y_1,z_1), (x_2,y_2,z_2), (x_3,y_3,z_3)\right)$ and associated 2D texture coordinates $\left((u_1,v_1), (u_2,v_2), (u_3,v_3)\right)$, Least Squares Conformal Mapping (LSCM)~\cite{10.1145/566654.566590} minimizes the angle distortion between corresponding triangle edges in 3D and 2D. 

It is implemented as a loss by measuring gradients of the texture mapping that maps points $(x,y)$ on the triangle $T$ to points $(u,v)$ on the texture through interpolation,
\begin{equation}
    \label{eq:lscm_energy}
    \mathcal{L}_{\text{LSCM}}(\mathcal{T}) = 
    \sum_{T \in \mathcal{T}} 
    \left(\left(\frac{\partial u}{\partial x} + \frac{\partial v}{\partial y}\right)^2
    + \left(\frac{\partial u}{\partial y} - \frac{\partial v}{\partial x}\right)^2
    \right) A,
\end{equation}
where 3D points $(x,y,z)$ are expressed in local triangle coordinates such that $z$ is aligned with the normal and is zero for points on $T$. 
The gradients are integrated over the triangle area $A$.
Following \cite{10.1145/566654.566590}, we compute the gradients as
\begin{equation*}
    \begin{pmatrix}
        \partial u / \partial x \\
        \partial u / \partial y
    \end{pmatrix}
    =
    \frac{0.5}{A} 
    \begin{pmatrix}
        y_2 - y_3 & y_3 - y_1 & y_1 - y_2 \\
        x_3 - x_2 & x_1 - x_3 & x_2 - x_1
    \end{pmatrix}\mkern-7mu
    \begin{pmatrix}
        u_1 \\
        u_2 \\
        u_3
    \end{pmatrix}\mkern-5mu.
\end{equation*}
The expressions for $\partial v / \partial x$ and $\partial v / \partial y$ are computed similarly by replacing $u$ with $v$.  

While the original LSCM optimizes the texture coordinates to ensure minimal distortion in texture mapping, we utilize the same energy to optimize the global $z$-vertex location, the depth, from the estimated texture coordinates.

\begin{figure*}[t!]
    \centering
    \includegraphics[trim={0 0 0 0},width=\textwidth]{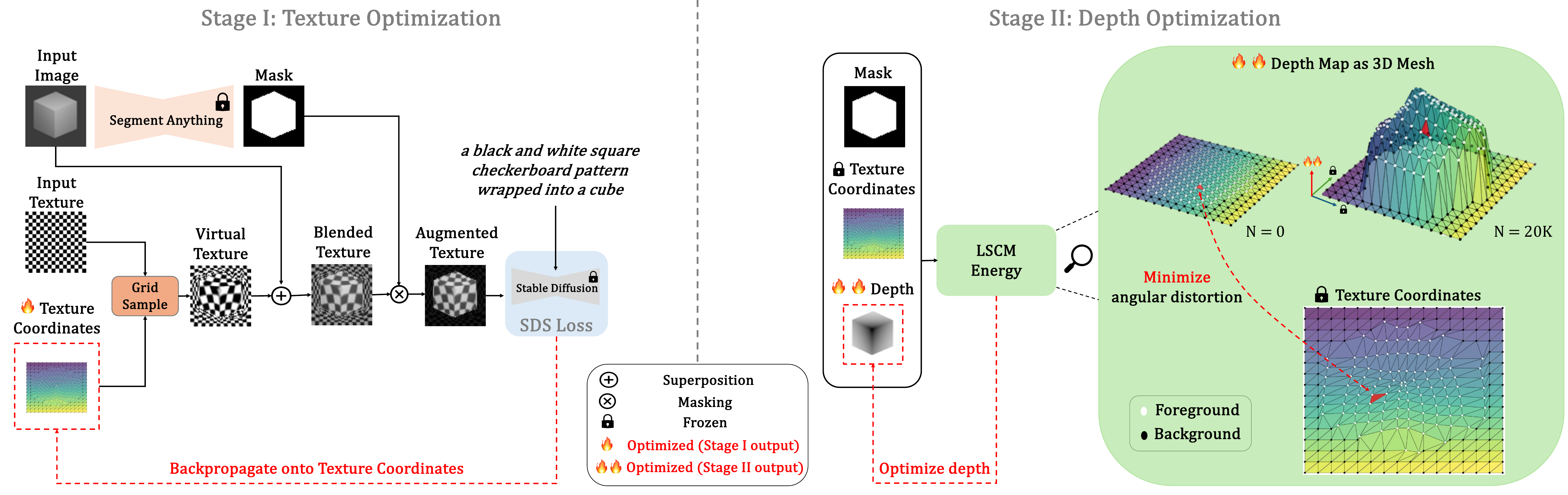}
    \caption{\textbf{DreamTexture overview.} %
    In Stage I, we use the SDS loss to faithfully augment the texture on the input image, producing texture coordinates that align the virtual texture with depth cues in the input image. In Stage II, we apply shape from virtual texture by viewing the depth map as the $z$-coordinates of a 3D mesh and optimize it using the LSCM energy, minimizing angular texture distortion.
    }
    \label{fig:overview}
\end{figure*}

\section{Method}
Given a single input image $\mI \in \mathbb{R}^{C \times W \times H}$, DreamTexture reconstructs a depth map $\mD \in \mathbb{R}^{W \times H}$ from monocular cues using the proposed analysis-by-augmentation method.

The key to our approach is the representation of the depth as a triangle mesh $\mathcal{T}$ with associated texture coordinates $\mW \in \mathbb{R}^{2 \times W \times H}$.
Figure~\ref{fig:overview} (right) shows how neighboring pixels in the depth map form faces with vertices $(i, j, \mD(i, j)) \in \mathbb{R}^{3}$ and associated texture coordinates $\mW(i,j)$. 
The mesh is special, with pixel positions $(i, j)$ frozen such that the mesh remains a regular grid. It lets us map one vertex to every pixel by representing depth and texture coordinates as tensors with dimensions equal to the input image. 
Figure~\ref{fig:overview} (left) shows how rendering the virtual texture boils down to sampling the flat input texture at the texture coordinates. 

This mesh representation lets us optimize a shared parametrization in two steps: In stage I, mesh vertex positions are frozen and only the texture coordinates are optimized to make the virtual texture look natural using SDS (Section~\ref{sec:preliminary}). %
Stage~II in turn optimizes the depth that parametrizes vertex coordinates while keeping the texture coordinates fixed. 
The triangle mesh representation is required such that the LSCM energy is applicable (Section~\ref{sec:preliminary}) to recover the depth that explains the frozen texture mapping---shape from virtual texture.

\paragraph{Assumptions} We exploit that texture and shading have an additive and multiplicative effect, which we model by affine blending. We furthermore use an orthographic projection model, which holds unless the object is close to the camera.

\paragraph{Initialization.} We estimate the foreground mask using the Segment Anything Model 2 (SAM 2)~\cite{kirillov2023segany}, use a checkerboard pattern as the regular texture, initialize the texture coordinates to preserve it and start with a spherical depth map unless otherwise stated in specific experiments.

\subsection{Augmentation Stage I - Texture Optimization}
\label{method:uv-deformation}

The goal is to deform a flat, regular texture $\mT \in \mathbb{R}^{C \times W \times H}$ by optimizing texture coordinates $\mW \in \mathbb{R}^{2 \times W \times H}$ until it aligns with $\mI$, yielding a virtually textured image $\tilde{\mT}$.

\paragraph{Texture augmentation.} 
We incorporate monocular depth cues from the input image $\mI$ by superimposing them with the virtual texture image $\tilde{\mT}$ through alpha blending  
\begin{equation}
    \tilde{\mI} = \mM \circ [\alpha \mI + (1-\alpha) \tilde{\mT}],
\end{equation}
where $\mM$ is a mask distinguishing foreground objects from the background in $\mI$, and $\alpha$ is a blend weight.

\paragraph{Stage I Loss.} The primary objective is to optimize $\mW$ with the SDS loss $\mathcal{L}_{\text{SDS}}(\tilde{\mI}, c)$ that evaluates the pre-trained image prior to quantify alignment of the augmentation with existing shape cues in the input image. Gradient descent is technically possible since the blending and the preceding texture sampling are differentiable.
However, without our subsequent extensions, direct optimization collapses, causing foldover in the texture coordinates.
We propose to optimize the texture coordinates hierarchically using a combination of SDS, $L_1$, and integrability~\cite{verbin2020toward} regularization,
\begin{equation}
\mathcal{L}_\text{tex}(\tilde{\mI}, c, \mW) = \mathcal{L}_{\text{SDS}}(\tilde{\mI}, c)  
    + \lambda_1 \mathcal{L}_{1}(\mW)  
    + \lambda_2 \mathcal{L}_{\text{int}}(\mW),
    \label{eq:loss}
\end{equation}
where $\lambda_1$ and $\lambda_2$ control the strength of regularization.

\paragraph{Relative texture coordinates.} To prevent degenerate deformations, we parametrize the texture coordinates $\mW = (\mW_u, \mW_v)$ through its spatial gradients $\mV = (\frac{\partial \mW_u}{\partial u}, \frac{\partial \mW_v}{\partial v}) $ as in~\cite{Shu_2018_ECCV}.
The absolute coordinates are reconstructed from $\mV$ via a spatial integration layer $\mW(\mV)$. 
Foldovers are prevented by enforcing non-negativity on these gradients with a Rectified Linear Unit (ReLU)~\cite{Agarap2018DeepLU}, smoothness is ensured with 
$\mathcal{L}_{1}(\mW(\mV)) = \| \mV \|_1$
, and local integrability with
\begin{align}
\mathcal{L}_{\text{int}}(\mW(\mV)) = \frac{1}{HW} \sum_{i,j}& \Big[ 
        \mV_u^{i, j+1} - \mV_u^{i+1, j+1} + \mV_u^{i, j} \nonumber \\
    & - \mV_u^{i+1, j} 
     + \mV_v^{i, j+1} + \mV_v^{i+1, j+1} \nonumber \\
    & - \mV_v^{i, j} - \mV_v^{i+1, j} 
    \Big]^2.
        \label{eq:warpfield losses}
\end{align}

\paragraph{Multiscale Optimization.} 
Rather than optimizing per-pixel spatial gradients $\mV$ directly, we adopt the multiscale approach from~\cite{barron2014shape}. Specifically, we optimize a set of $N$ scales $\{\mV^{(j)}\}_{j=0}^{N-1}$, where each $\mV^{(j)}$ has dimensions $W/2^j \times H/2^j$ and represents a coarse-to-fine hierarchy. The full-resolution gradients are then reconstructed as
\begin{equation}
    \mV = \mathcal{G}^T(\mV^{(0)} \circ \cdots \circ \mV^{(N-1)}).
\end{equation}

To construct the Gaussian pyramid $\mathcal{G}$, we apply a 2D kernel $k^T \cdot k$, where $k = m \cdot \frac{1}{16}[1, 4, 6, 4, 1]$, and $m = 1.4$, as in~\cite{barron2014shape}. This hierarchical representation improves texture deformations by progressively refining coarse structures before fine-grained details.

\paragraph{Text Prompt.} 
To encourage the diffusion model to perceive the augmentation as a checkerboard texture, we set the input text prompt to \textit{a black and white square checkerboard pattern wrapped into a \{object\}}. The object description could also be determined automatically using large language models (LLMs) such as LaMa~\cite{Touvron2023LLaMAOA}.

\subsection{Analysis Stage II - Depth Optimization}
\label{method:depth-recon}

This stage can be seen as performing shape from texture on the virtual texture optimized from Stage I. Opposed to classical SfT, our texture coordinates provide absolute texture position, which eases reconstruction. It lets us apply the LSCM energy to optimize depth such that 
the texture appears undistorted on the surface of the 3D mesh $\cT$.

The LSCM energy is typically used to find a conformal, i.e., deformation preserving, texture mapping given a static mesh. Here we keep texture coordinates frozen and optimize vertex positions instead. Figure~\ref{fig:overview} (right) gives an example. When initialized with zero depth, all 3D triangles have the same size and angles but some triangles in texture coordinates are stretched. During the optimization with the LSCM energy, the corresponding depth values are altered to form a correspondingly stretched triangle in 3D. To improve stability we minimize the LSCM energy together with an $L_1$ regularization on depth gradients,
\begin{equation}
    \mathcal{L}_{\text{depth}}(\mathcal{T}, \mD, \mW) =  \mathcal{L}_{\text{LSCM}}(\mathcal{T}, \mW) + \frac{\lambda_3}{hw} \sum_{i,j} \left| \Delta \mD^{i,j} \right|
    \label{eq:depth}
\end{equation}
with respect to $\mD(i,j)$, keeping only triangles formed by foreground pixels in the mask $\mM$ for $\mathcal{L}_{\text{LSCM}}$. The parameter $\lambda_3$ is the regularization weight.

The outcome is a 3D mesh that has a regular texture in 3D while its projection augments the input image, such that the virtual texture appears like a natural texture that follows the geometry in the image---analysis by augmentation.

\begin{table*}[t!]
    \centering
    \begin{minipage}[t]{0.47\textwidth}
    \renewcommand{\arraystretch}{0}  %
    \renewcommand{\tabcolsep}{0pt}
    \begin{tabularx}{\textwidth}{c *{6}{Y}}
        & Input & 
         Recon. & 
         Recon. & 
         Render & 
         Render & 
         Render \\
         & & 
         Depth &
         Normal &
         View 1 & 
         View 2 & 
         View 3 \vspace{0.2cm}\\
        \rotatebox{90}{\hspace{0.5cm}DF} &
        \multicolumn{6}{c}{\includegraphics[width=0.95\textwidth]{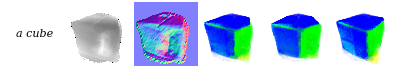}}\\
        \rotatebox{90}{\hspace{0.5cm}RF} &
        \multicolumn{6}{c}{\includegraphics[width=0.95\textwidth]{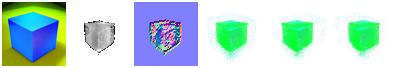}}\\
        \rotatebox{90}{\hspace{0.3cm}\textbf{Ours}} &
        \multicolumn{6}{c}{\includegraphics[width=0.95\textwidth]{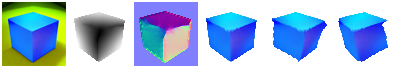}} \vspace{0.2cm}\\
        \rotatebox{90}{\hspace{0.5cm}DF} &
        \multicolumn{6}{c}{\includegraphics[width=0.95\textwidth]{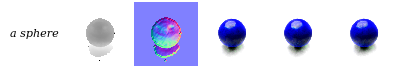}}\\
        \rotatebox{90}{\hspace{0.5cm}RF} &
        \multicolumn{6}{c}{\includegraphics[width=0.95\textwidth]{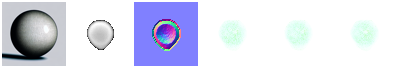}}\\
        \rotatebox{90}{\hspace{0.3cm}\textbf{Ours}} &
        \multicolumn{6}{c}{\includegraphics[width=0.95\textwidth]{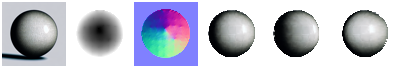}}\vspace{0.2cm}\\
        \rotatebox{90}{\hspace{0.5cm}DF} &
        \multicolumn{6}{c}{\includegraphics[width=0.95\textwidth]{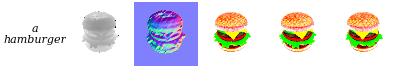}} \\
        \rotatebox{90}{\hspace{0.5cm}RF} &
        \multicolumn{6}{c}{\includegraphics[width=0.95\textwidth]{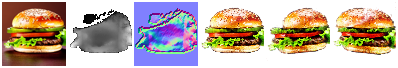}}\\
        \rotatebox{90}{\hspace{0.3cm}\textbf{Ours}} &
        \multicolumn{6}{c}{\includegraphics[width=0.95\textwidth]{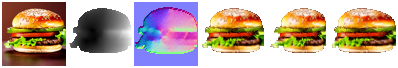}}\\
    \end{tabularx}
    \end{minipage}
    \hfill
    \begin{minipage}[t]{0.47\textwidth}
    \renewcommand{\arraystretch}{0}  %
    \renewcommand{\tabcolsep}{0pt}
    \begin{tabularx}{\textwidth}{c *{6}{Y}}
        & Input & 
         Recon. & 
         Recon. & 
         Render & 
         Render & 
         Render \\
         & & 
         Depth &
         Normal &
         View 1 & 
         View 2 & 
         View 3 \vspace{0.2cm}\\
        \rotatebox{90}{\hspace{0.5cm}DF} &
        \multicolumn{6}{c}{\includegraphics[width=0.95\textwidth]{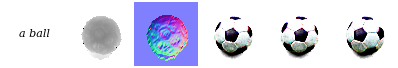}}\\
        \rotatebox{90}{\hspace{0.5cm}RF} &
        \multicolumn{6}{c}{\includegraphics[width=0.95\textwidth]{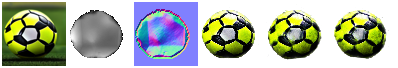}}\\
        \rotatebox{90}{\hspace{0.3cm}\textbf{Ours}} &
        \multicolumn{6}{c}{\includegraphics[width=0.95\textwidth]{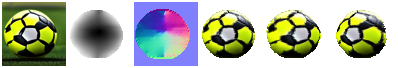}} \vspace{0.2cm}\\
        \rotatebox{90}{\hspace{0.5cm}DF} &
        \multicolumn{6}{c}{\includegraphics[width=0.95\textwidth]{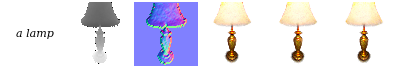}}\\
        \rotatebox{90}{\hspace{0.5cm}RF} &
        \multicolumn{6}{c}{\includegraphics[width=0.95\textwidth]{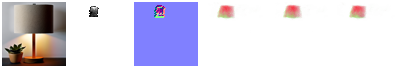}}\\
        \rotatebox{90}{\hspace{0.3cm}\textbf{Ours}} &
        \multicolumn{6}{c}{\includegraphics[width=0.95\textwidth]{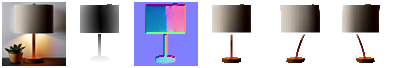}}\vspace{0.2cm}\\
        \rotatebox{90}{\hspace{0.5cm}DF} &
        \multicolumn{6}{c}{\includegraphics[width=0.95\textwidth]{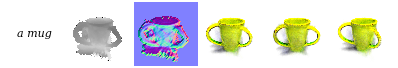}} \\
        \rotatebox{90}{\hspace{0.5cm}RF} &
        \multicolumn{6}{c}{\includegraphics[width=0.95\textwidth]{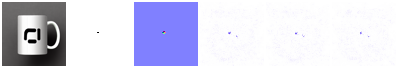}}\\
        \rotatebox{90}{\hspace{0.3cm}\textbf{Ours}} &
        \multicolumn{6}{c}{\includegraphics[width=0.95\textwidth]{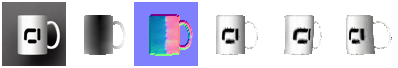}}\\
    \end{tabularx}
    \end{minipage}
    \captionof{figure}{\textbf{Text-to-3D comparison.} Compared to RealFusion (RF) and DreamFusion (DF), our method achieves more accurate reconstructions, particularly for objects that lack prominent visual features. We use the same text prompt as DreamFusion to generate input images with Stable Diffusion. To render novel views in our approach, we resample the input image using the inverse texture coordinates
    and render using PyTorch3D with orthographic projection (elevation $=0^{\circ}$) and azimuth angles of $0^{\circ}$ (view 1), $-10^{\circ}$ (view 2), and $+10^{\circ}$ (view 3).}
    \label{fig:text-to-3d}
\end{table*}

\section{Experiments}

We conduct extensive experiments to evaluate the effectiveness of DreamTexture and the associated
 analysis-by-augmentation principle for depth reconstruction from a single image and demonstrate its computational efficiency. Figure~\ref{fig:text-to-3d} shows the main results, additional ones are provided in the appendix.

\paragraph{Datasets.} For qualitative and quantitative evaluation of our proposed approach, we construct a synthetic dataset, \emph{PrimitiveShapesX}, consisting of four basic primitives (cube, pyramid, cylinder, and sphere) in Blender. Each primitive is rendered from a single camera view, and we record both the rendered image and the corresponding depth map. The simplicity of these shapes is chosen to evaluate flat, curved, and cornered objects in isolation. To explore the influence of different monocular depth cues, we render each primitive under three conditions: without any texture (shaded only), with a regular (repetitive) texture, and with a natural (wood) texture.         

We also evaluate qualitatively on dreamed-up images and pictures of real objects without ground truth.

\paragraph{Metrics.}

We use Mean Squared Error (MSE) to measure the discrepancy between the ground truth and reconstructed depth maps, computing it only within the foreground region. %
To overcome the scale ambiguity in monocular reconstruction, we normalize predictions to zero mean and unit variance, then rescale it to match the ground truth statistics. For text-to-3D evaluation without a ground truth depth map, we apply min-max normalization within the foreground. 

\paragraph{Baselines.} We evaluate our approach against DreamFusion (DF)~\cite{poole2022dreamfusion} and RealFusion (RF)~\cite{MelasKyriazi2023RealFusion3R}, both of which rely exclusively on multi-view cues without using any pre-trained depth or normal estimators. As the official implementation of DreamFusion is not publicly available, we utilize a widely adopted open-source reimplementation (\url{https://github.com/chinhsuanwu/dreamfusionacc}). To ensure a fair comparison, we employ for all models the same Stable Diffusion model for guidance and render images at a resolution of $64 \times 64$ as in the original DreamFusion. All other hyperparameters for both methods follow the default values specified in their respective example configurations.

We focus on the unsupervised reconstruction regime and hence exclude related methods that utilize pre-trained monocular depth estimators~\cite{xu2023neurallift360, tang2023makeit3d}.

\paragraph{Implementation Details.} We conduct our experiments following the original setup of DreamFusion, with a resolution of \(64 \times 64\) for both the depth map and corresponding texture coordinates. For SDS guidance, we utilize Stable Diffusion v1.5~\cite{Rombach2021HighResolutionIS}. Since Stable Diffusion operates at a resolution of \(512 \times 512\), we upscale the augmented texture to \(512 \times 512\) before feeding it into the diffusion model. The texture is blended onto the input image using a factor of \(\alpha = 0.5\).  

For optimization, we adopt the AdamW optimizer with a learning rate of \(5 \times 10^{-5}\) for the texture coordinates and \(10^{-3}\) for the depth map. The weighting coefficients are set to \(\lambda_{1} = 10\), \(\lambda_{2} = 10^{4}\), and \(\lambda_{3} = 0.01\). Stage I is optimized for 100K iterations, with snapshots taken every 10K iterations. Stage II is optimized for 20K iterations for each. Among these, we select the smoothest depth map as the final output i.e. the depth map with the lowest \( \Delta \mD(i,j) \).  

All experiments are conducted on a single NVIDIA A40 GPU. To compute surface normals, we take the cross-product of the numerical horizontal and vertical gradients of the depth map. Further implementation details are provided in Appendix~\ref{app:implementationdetails}.

\begin{table}[t]
    \centering
    \renewcommand{\arraystretch}{0}  %
    \renewcommand{\tabcolsep}{0pt}
    \begin{tabularx}{0.98\linewidth}{Y*{6}{Y}}
    & & \multicolumn{2}{c}{\textcolor[RGB]{78, 167, 46}{\emph{Stage II Output}}} & \multicolumn{2}{c}{\textcolor[RGB]{233, 115, 50}{\emph{Stage I Output}}}\vspace{0.2cm}\\
        Input & 
        GT & 
        Recon. & 
        Recon. & 
        Texture & 
        Augmented\\
        Image &
        Depth & 
        Depth & 
        Normal & 
        Coord. & 
        Texture\vspace{0.2cm}\\%
        \multicolumn{6}{c}{\includegraphics[trim={0 0 0 0}, clip, width=0.98\linewidth]{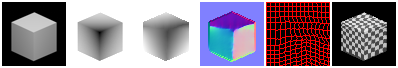}}\\
        \multicolumn{6}{c}{\includegraphics[trim={0 0 0 0}, clip, width=0.98\linewidth]{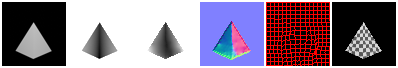}}\\
        \multicolumn{6}{c}{\includegraphics[trim={0 0 0 0},clip, width=0.98\linewidth]{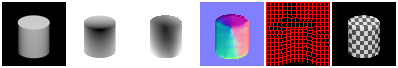}}\\
        \multicolumn{6}{c}{\includegraphics[trim={0 0 0 0},clip, width=0.98\linewidth]{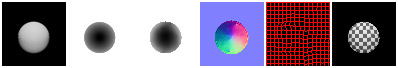}}\\
    \end{tabularx}
    \captionof{figure}{\textbf{Qualitative evaluation} on \emph{PrimitiveShapesX}. Our method reconstructs depth maps with high fidelity, producing accurate surface normals for smooth and sharp-edged objects.}
    \label{fig:qualitative}
\end{table}

\subsection{Qualitative Evaluation}
To highlight the effectiveness of our approach, we first evaluate it on the \emph{PrimitiveShapesX} dataset, as shown in Figure~\ref{fig:qualitative}. Evaluating on basic primitives allows us to assess whether the method can accurately model curved surfaces, flat planes, and sharp edges. Our results demonstrate that the depth maps are reconstructed with high fidelity, while the estimated surface normals are both accurate and smooth. Additionally, the augmented texture image exhibits correct texture mapping onto the object with minimal distortion. This suggests that the SDS loss effectively models the texture coordinates to ensure accurate texture augmentation.

\begin{table}[t!]
    \centering
    \caption{\textbf{Quantitative Comparison with RealFusion (RF)} on the naturally textured \emph{PrimitiveShapesX} dataset. We outperform RealFusion by a significant margin due to better robustness.}
    \label{table:realfusioncomp}
    \resizebox{0.7\linewidth}{!}{
        \begin{tabular}{lcccc}
            \toprule
            \multirow{2}{*}{Object} & \multicolumn{2}{c}{Depth MSE \textdownarrow} & \multicolumn{2}{c}{Normal MSE \textdownarrow}\\
            \cmidrule(lr){2-3}
            \cmidrule(lr){4-5}
            & RF & \textbf{Ours} & RF & \textbf{Ours}\\
            \midrule
            Cube & 0.0706 &  \textbf{0.0071} & 0.5618 & \textbf{0.0673}\\
            Sphere & 0.0969 & \textbf{0.0062} &  0.6729 & \textbf{0.0291}\\
            Pyramid & 0.0302 &  \textbf{0.0088} &  0.4007 & \textbf{0.0675}\\
            Cylinder & 0.0774 &  \textbf{0.0255} & 0.5544 & \textbf{0.1097}\\
            \bottomrule
        \end{tabular}
    }
\end{table}

\subsection{Quantitative Evaluation}
Table~\ref{table:realfusioncomp} presents a quantitative comparison of our approach against RealFusion (RF) on the naturally textured \emph{PrimitiveShapesX} dataset. We use naturally textured objects as input images, as Stable Diffusion is trained primarily on real-world images. Our results demonstrate a substantial performance improvement over RealFusion. 

\begin{table}[t!]
    \centering
    \caption{\textbf{Efficiency Comparison} in terms of training parameters and training time, measured on a single Nvidia A40 GPU. Our method has a significantly lower parameter count and achieves a shorter training time per iteration. \textbf{Ours (100K)} runs for 100K iterations, and \textbf{Ours (10K)} runs for 10K in Stage I.}
    \label{table:dreamfusioncomp}
    \resizebox{\linewidth}{!}{
        \begin{tabular}{lcccccc}
            \toprule
            \multirow{2}{*}{Method} & \multicolumn{3}{c}{Trainable Parameters \textdownarrow} & \multicolumn{3}{c}{Training Time (s) \textdownarrow}\\
            \cmidrule(lr){2-4}
            \cmidrule(lr){5-7}
            & Stage I & Stage II & Total & Stage I & Stage II & Total\\
            \midrule
            DF & 12604016 & -  & 12604016 & 3126.2 & - & 3126.2\\
            RF & 37946112 & 1806983 & 39753095 & 3384,3 & 1470.7 & 4855.0\\
            \textbf{Ours (100K)} & 10922 &  4096 & \textbf{15018} & 10679.2 & 1408.7 & 12087.9\\
            \textbf{Ours (10K)} & 10922 &  4096 & \textbf{15018} & 1070.0 & 140.8 & \textbf{1210.9}\\
            \bottomrule
        \end{tabular}
    }
\end{table}

Furthermore, we compare our method against DreamFusion and RealFusion in terms of the number of trainable parameters and training time per iteration, as summarized in Table~\ref{table:dreamfusioncomp}. Notably, the implementations of DreamFusion and RealFusion leverage Instant Neural Graphics Primitives~\cite{Mller2022InstantNG}, a highly efficient implementation of their NeRF model. This differs from the original DreamFusion implementation, which employs a standard NeRF. Nevertheless, our method has significantly fewer parameters and requires a shorter training time for the same number of iterations.

\subsection{Text-to-3D Evaluation}
We compare our text-to-3D generation results against the state-of-the-art approaches, DreamFusion (DF)~\cite{poole2022dreamfusion} and RealFusion (RF)~\cite{MelasKyriazi2023RealFusion3R}, in Figure~\ref{fig:text-to-3d}. To ensure a fair comparison, we use the same text prompt as DreamFusion to generate the input image for RealFusion and our method.

Our results demonstrate that DreamFusion struggles to accurately reconstruct a cube, instead generating a five-sided die. Additionally, it exhibits artifacts around shading highlights, particularly on surfaces such as spheres. RealFusion, on the other hand, performs poorly on objects like lamps and mugs. In contrast, our method successfully reconstructs simple objects. However, for more complex objects, such as a hamburger, we observe missing details in the reconstruction. We attribute this limitation to the low resolution of the texture coordinates and the reliance only on virtual texture cues. Their complementary strengths suggest future work combining additional monocular cues or multi-view information.

We provide further analysis of RealFusion in Appendix~\ref{app:realfusion}. 

\begin{table*}[t!]
    \centering
    \renewcommand{\arraystretch}{1}  %
    \centering
    \caption{\textbf{Quantitative ablation study on the effect of input image cues.} We study the effect of four different cues in the input image: silhouette,  shading, regular texture, and natural texture. The input with shading and silhouette cues achieves the lowest errors.
    }
    \label{table:cue}
    \resizebox{\linewidth}{!}{
        \begin{tabular}{lcccccccccc}
            \toprule
            \multirow{2}{*}{Cue} & 
            \multicolumn{5}{c}{Depth MSE \textdownarrow} &
            \multicolumn{5}{c}{Normal MSE \textdownarrow}\\
            \cmidrule(lr){2-6}
            \cmidrule(lr){7-11} 
            & Cube & Sphere & Pyramid & Cylinder & Avg.
            & Cube & Sphere & Pyramid & Cylinder & Avg.\\
            \midrule
            Silhouette & 0.0080 & 0.0067 & 0.0240 & 0.0175 & 0.0140
            & 0.0996 & 0.0334 & 0.2122 & 0.1230 & 0.1170\\
            Silhouette + Shading & \textbf{0.0070} & \textbf{0.0046} & \textbf{0.0079} & 0.0162 & \textbf{0.0089}
            & \textbf{0.0629} & \textbf{0.0177} & 0.0858 & 0.0905 & \textbf{0.0642}\\
            Silhouette + Shading + Regular Tex & 0.0092 & 0.0053 & 0.0107 & \textbf{0.0136} & 0.0358
            & 0.1545 & 0.0195 & 0.1074 & \textbf{0.0655} & 0.0867 \\
            Silhouette + Shading + Natural Tex & 0.0071 & 0.0062 & 0.0088 & 0.0255 & 0.0119 
            & 0.0673 & 0.0291 & \textbf{0.0675} & 0.1097 & 0.0684 \\
            \bottomrule
        \end{tabular}
    }
\end{table*}

\begin{table}[t!]
    \centering
    \renewcommand{\arraystretch}{0}  %
    \renewcommand{\tabcolsep}{0pt}
    \begin{tabularx}{0.9\linewidth}{c@{\hspace{0.15cm}}c@{\hspace{0.2cm}}Y*{5}{Y}}
        &
        &
        Input &
        GT &
        Recon. & 
        Recon. \\
        &
        &
        Image &
        Depth &
        Depth &
        Normal \\
        \rotatebox{90} {\hspace{0.7cm}S} &
        &
        \multicolumn{4}{c}{\includegraphics[trim={0 0 4.67cm 0}, clip, width=0.8\linewidth]{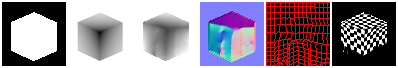}} \\
        \rotatebox{90}{\hspace{0.35cm}S + SH} &
        &
        \multicolumn{4}{c}{\includegraphics[trim={0 0 4.67cm 0}, clip, width=0.8\linewidth]{images/image-to-3d/cube.png}} \\
        \rotatebox{90}{\hspace{0.35cm}S + SH} &
        \rotatebox{90}{\hspace{0.35cm}+ RT} &
        \multicolumn{4}{c}{\includegraphics[trim={0 0 4.67cm 0}, clip, width=0.8\linewidth]{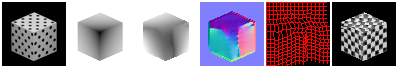}} \\
        \rotatebox{90}{\hspace{0.35cm}S + SH} &
        \rotatebox{90}{\hspace{0.35cm}+ NT} &
        \multicolumn{4}{c}{\includegraphics[trim={0 0 4.67cm 0}, clip, width=0.8\linewidth]{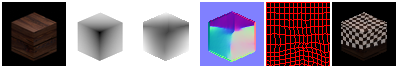}} \\
    \end{tabularx}
    \captionof{figure}{\textbf{Qualitative ablation} study on the effect of combining four different input image cues: 
    silhouette (S), shading (SH), regular texture (RT), and natural texture (NT). The input with silhouette and shading (S + SH) cues yields the most accurate and smooth geometry, confirming the quantitative results.}
    \label{fig:cue}
\end{table}

\subsection{Ablation Study}

Our virtual shape-from-texture approach could only be realized through a set of design choices and extensions. The extensive ablation study in Appendix~\ref{app:ablation} isolates and quantifies the effect of each.
In Figure~\ref{fig:cue} and Table~\ref{table:cue}, we analyze the impact of different shape cues in the input image. We evaluate four different input complexities, with an increasing number of depth cues: (0) silhouette, (1) shading only, (2) shading with regular texture (circular patterns), and (3) shading with natural texture (wood texture).  Our results indicate that while a silhouette is sufficient to capture the rough geometry, the resulting surface lacks details. Although adding texture helps to smooth the geometry, it introduces some artifacts around the edges. In this setting, a textureless scene (shading only) provides the strongest cues for recovering accurate geometry.

\begin{table}[t]
    \centering
    \renewcommand{\arraystretch}{0}  %
    \renewcommand{\tabcolsep}{0pt}
    \begin{tabularx}{0.99\linewidth}{Y*{6}{Y}}
        Input & 
         Recon. & 
         Recon. & 
         Render & 
         Render & 
         Render \\
         & 
         Depth &
         Normal &
         View 1 & 
         View 2 & 
         View 3 \vspace{0.2cm}\\
        \multicolumn{6}{c}{\includegraphics[trim={0 0 0cm 0}, clip, width=0.99\linewidth]{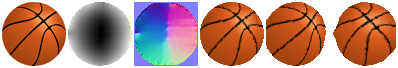}} \\
        \multicolumn{6}{c}{\includegraphics[trim={0 0 0cm 0}, clip, width=0.99\linewidth]{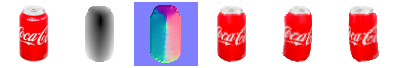}} \\
    \end{tabularx}
    \captionof{figure}{\textbf{In-the-wild reconstruction.} DreamTexture reconstructs 3D objects from single-view in-the-wild images with high fidelity.}
    \label{fig:in-the-wild}
\end{table}

\begin{table}[t]
    \centering
    \renewcommand{\arraystretch}{0}  %
    \renewcommand{\tabcolsep}{0pt}
    \begin{tabularx}{0.5\linewidth}{Y*{3}{Y}}
        Input & 
        Recon. & 
        Recon. \\
        Image &
        Depth &
        Normal \\
        \multicolumn{3}{c}{\includegraphics[trim={0 0 4.67cm 0}, clip, width=0.5\linewidth]{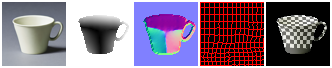}} \\
        \multicolumn{3}{c}{\includegraphics[trim={0 0 4.67cm 0}, clip, width=0.5\linewidth]{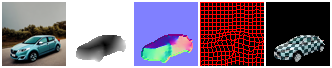}} \\
    \end{tabularx}
    \captionof{figure}{\textbf{Limitations.} Our method struggles with complex objects such as cars and fails to resolve concave and convex ambiguities, treating objects like cups as convex instead of concave.}
    \label{fig:limitation}
\end{table}

\subsection{Reconstruction in the Wild}
We demonstrate the application of our method for reconstructing in-the-wild images from a single view. In Figure~\ref{fig:in-the-wild}, we present images taken from the internet, along with the depth and normal maps reconstructed by our pipeline. These results show that our approach is capable of estimating depth from single-view in-the-wild images.

\subsection{Limitations and Future Work} 

As our method introduces a new paradigm for object reconstruction, this first realization has several limitations and areas for improvement. Our depth map reconstructs only the visible part of the object but it could be combined with virtual multi-view reconstruction.
In Figure~\ref{fig:limitation}, we present representative failure cases. Our approach struggles with complex objects, such as the car. Future work could improve by partitioning the image into parts and by augmenting additional monocular cues, such as shading and contours.
These could also overcome the concave and convex ambiguities visible in the cup example, demonstrated for classical SfT in combination with shape from shading~\cite{white2006combining}.

\section{Conclusion}

Making machines learn by themselves remains a key challenge in AI. Towards this goal, our shape-from-virtual-texture implementation DreamTexture provides an efficient unsupervised 3D reconstruction method. 
The underlying analysis-by-augmentation principle applies more broadly, to shape cues beyond texture, and offers new ways for leveraging pre-trained generative models for reconstruction, paving the way for unsupervised learning at scale.

\section*{Acknowledgments}

We thank all the members of Visual AI for Extended Reality lab at Bielefeld University including Bianca Schröder, Jerin Philip and Sebastian Dawid for providing crucial feedback and participating in helpful discussions.
 
{
    \small
    \bibliographystyle{ieeenat_fullname}
    \bibliography{main}
}

\clearpage
\setcounter{page}{1}
\maketitlesupplementary
\appendix

This document provides additional details and results that complement the main document.

\section{Implementation Details}
\label{app:implementationdetails}
We realize the Score Distillation Sampling (SDS) that is defined in  Equation~\ref{eq:loss} through its gradient $\nabla_{\theta} \mathcal{L}_{\text{SDS}}(\mX, c)$
as an MSE loss,
\begin{equation}
    \mathcal{L}_{\text{SDS}}(\mathbf{X}, c) = \mathop{\mathbb{E}}_{\epsilon, t} \left[\frac{1}{2} w(t) \left( \mathbf{X} - \mZ \right)^2\right],
\end{equation}
where $\mZ = \left( \mathbf{X} - (\epsilon_{\phi}(\mathbf{X}_t; c, t) - \epsilon) \right)$ is taken as a constant,
\( \epsilon \sim \mathcal{N}(\mathbf{0}, \mathbf{I}) \), and \( \mX \) represents the rendering from a differentiable representation with parameters \( \theta \), and \( \epsilon_{\phi}(\mathbf{X}_t; c, t) \) is the predicted noise from the diffusion prior, parameterized by \( \phi \), given the noisy image \( \mX_t \), text embedding \( c \), and noise level \( t \).  

We implement the discrete Laplacian operator \( \Delta \mD(i,j) \) at pixel location \( (i, j) \) in Equation~\ref{eq:depth} as:
\begin{equation}
\begin{aligned}
    \Delta \mD^{i,j} = \mD^{i+1,j} + \mD^{i-1,j}
                + \mD^{i,j+1} 
                + \mD^{i,j-1} - 4\mD^{i,j}.
\end{aligned}
\end{equation}

\section{Motivation for the LSCM Energy}
To align the virtual textured image with monocular cues from the input image, the SDS loss induces distortions in the texture coordinates field. This effect is illustrated in Figure~\ref{fig:lscmvssds}. Specifically, to match the sphere in the input image, the field exhibits highly stretched triangles near the boundaries, while those near the center appear more compressed or less stretched. This suggests that large stretches in the texture coordinates field correspond to curved or deeper regions in 3D space, whereas smaller stretches indicate flatter areas.  

A similar effect is observed when generating the texture coordinates field using the Least Squares Conformal Map (LSCM). In this case, triangles near the boundaries in the texture space appear stretched, whereas those closer to the center remain compressed. Therefore, we use the same LSCM energy to infer the 3D shape treating the depth map as a 3D mesh from the optimized texture coordinates.

\begin{table}[t!]
    \centering
    \renewcommand{\arraystretch}{1}  %
    \begin{tabular}{cc}
        \hspace{0.5cm} Input Image & 
        \hspace{0cm} Texture Coordinates \\
        \multicolumn{2}{c}{\includegraphics[trim={0 0 0 0}, clip, width=0.7\columnwidth]{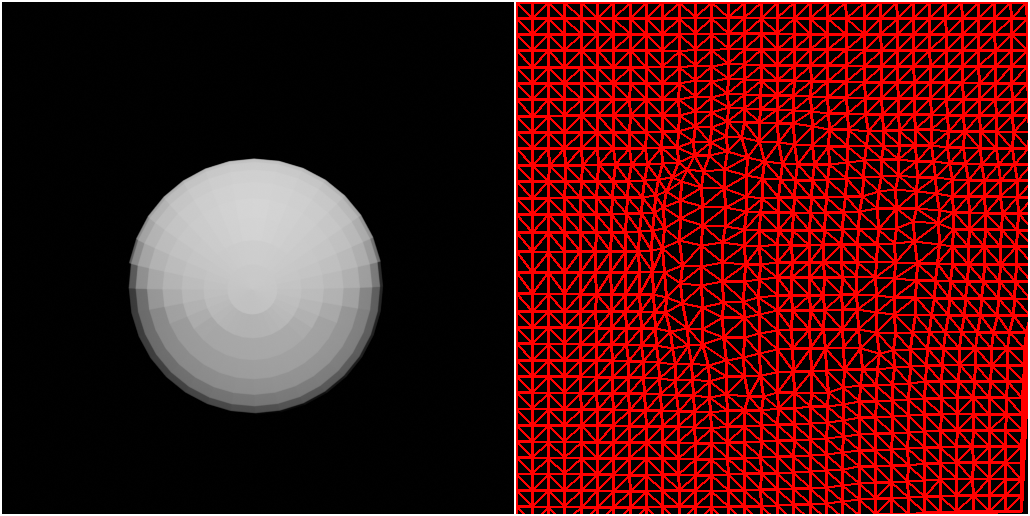}} \\
        \hspace{0.5cm} GT Depth & 
        \hspace{0.0cm} LSCM Texture Coordinates \\
        \multicolumn{2}{c}{\includegraphics[trim={0 0 0 0}, clip, width=0.7\columnwidth]{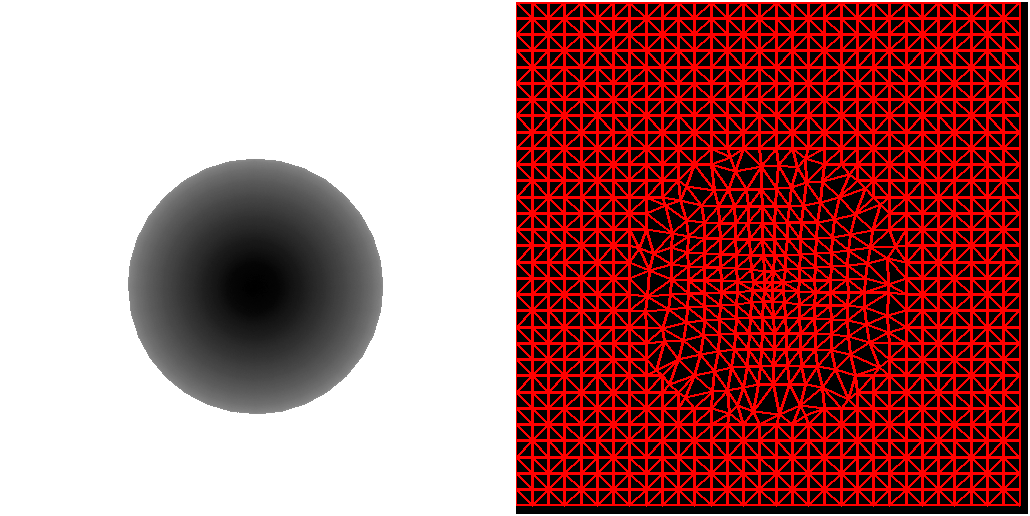}} \\
    \end{tabular}
    \captionof{figure}{\textbf{Effect of SDS on texture coordinates for a sphere.} SDS deforms the texture coordinate field by stretching triangles near the poles and equator to better align the texture with the input image. A similar effect is observed when applying LSCM to the corresponding ground-truth depth of the sphere, treating the depth map as a 3D mesh. This demonstrates that LSCM energy can be utilized to optimize a depth map to capture the distortions introduced by SDS in the 2D texture coordinates. \textit{Electronic zoom-in recommended.}}
    \label{fig:lscmvssds}
\end{table}

\begin{table}[t!]
    \centering
    \renewcommand{\arraystretch}{1}  %
    \begin{tabular}{cc}
        \hspace{0.9cm} GT Depth
        &
        \hspace{0.5cm} Recovered Depth
        \\
        &
        \hspace{0.5cm} (MSE: $2.152 \times 10^{-5}$)
        \\
        \multicolumn{2}{c}{\includegraphics[trim={0 0.35cm 0 0},clip, width=0.4\textwidth]{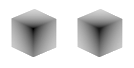}}
    \end{tabular}
    \captionof{figure}{\textbf{Experiment validating bidirectional optimization in LSCM via gradient descent.} We first optimize the texture coordinates coordinates using the ground-truth (GT) depth map. Then, starting from an initial spherical depth, we optimize the depth map using the previously optimized texture coordinates. The low MSE of the reconstructed depth verifies the feasibility of both forward and backward optimization in LSCM.}
    \label{fig:depth-rec-verification}
\end{table}

\begin{table*}[t!]
    \centering
    \renewcommand{\arraystretch}{1}  %
    \centering
    \caption{\textbf{Quantitative ablation study for the changes in loss function weights, blend weight, and texture coordinates parametrization.}}
    \label{table:ablation2}
    \resizebox{\linewidth}{!}{
        \begin{tabular}{lcccccccccc}
            \toprule
            \multirow{2}{*}{Cue} & 
            \multicolumn{5}{c}{Depth MSE \textdownarrow} &
            \multicolumn{5}{c}{Normal MSE \textdownarrow}\\
            \cmidrule(lr){2-6}
            \cmidrule(lr){7-11} 
            & Cube & Sphere & Pyramid & Cylinder & Avg.
            & Cube & Sphere & Pyramid & Cylinder & Avg.\\
            \midrule
            Ours & 0.0070 & \textbf{0.0046} & 0.0079 & 0.0162 & \textbf{0.0089}
            & 0.0629 & \textbf{0.0177} & 0.0858 & 0.0905 & \textbf{0.0642}\\

            \dots\, $\lambda_1=0$ & 0.0137 & 0.0144 & \textbf{0.0074} & \textbf{0.0110} & 0.0116
            & 0.0786 & 0.0938 & \textbf{0.0771} & 0.1049 & 0.0886\\
            
            \dots\, $\lambda_2=0$ & 0.0128 & 0.0083 & 0.0107 & 0.0131 & 0.0112
            
            & 0.0765 & 0.0497 & 0.1139 & 0.1015 & 0.0854\\
            
            \dots\, $\lambda_3=0$ & 0.0087 & 0.0169 & 0.0315 & 0.0263 & 0.0208
            
            & 0.0954 & 0.1741 & 0.2878 & 0.1506 & 0.1770\\
            
            \dots\, $\alpha=0.3$ & \textbf{0.0048} & 0.0108 & 0.0158 & 0.0166 & 0.0120
            
            & 0.0613 & 0.0639 & 0.1129 & 0.1102 & 0.0871\\
            
            \dots\, $\alpha=0.7$ & 0.0057 & 0.0081 & 0.0205 & 0.0176 & 0.0130
            
            & 0.0646 & 0.0367 & 0.1503 & \textbf{0.0798} & 0.0829\\
        
            \dots\, w/o multi-scale texture coordinates & 0.0088 & 0.0055 & 0.0120 & 0.0178 & 0.0110
            
            & \textbf{0.0567} & 0.0256 & 0.1051 & 0.0806 & 0.0670\\
            \bottomrule
        \end{tabular}
    }
\end{table*}

\begin{table}[t]
    \centering
    \renewcommand{\arraystretch}{0}  %
    \renewcommand{\tabcolsep}{0pt}
    \begin{tabularx}{0.99\linewidth}{c@{\hspace{0.15cm}}c@{\hspace{0.05cm}}Y*{6}{Y}}
        &
        &
        Input & 
        GT & 
        Recon. & 
        Recon. & 
        Texture \\
        &
        &
        Image &
        Depth & 
        Depth & 
        Normal & 
        \hspace{0.1cm}Coord. \vspace{0.2cm}\\%

        \rotatebox{90}{\hspace{0.35cm}Ours} & &
        \multicolumn{5}{c}{\includegraphics[trim={0 0 2.4cm 0cm}, clip, width=0.89\linewidth]{images/image-to-3d/cube.png}}\\
        
        \rotatebox{90}{\hspace{0.3cm}$\lambda_1=0$} & &
        \multicolumn{5}{c}{\includegraphics[trim={0 0 2.4cm 0cm}, clip, width=0.89\linewidth]{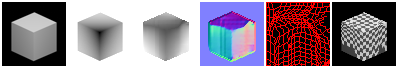}}\\

        \rotatebox{90}{\hspace{0.3cm}$\lambda_2=0$} & &
        \multicolumn{5}{c}{\includegraphics[trim={0 0 2.4cm 0cm}, clip, width=0.89\linewidth]{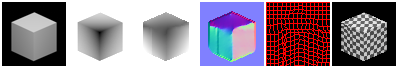}}\\

        \rotatebox{90}{\hspace{0.3cm}$\lambda_3=0$} & &
        \multicolumn{5}{c}{\includegraphics[trim={0 0 2.4cm 0cm},clip, width=0.89\linewidth]{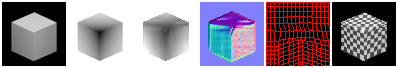}}\\

        \rotatebox{90}{\hspace{0.2cm}$\alpha=0.3$} & &
        \multicolumn{5}{c}{\includegraphics[trim={0 0 2.4cm 0cm},clip, width=0.89\linewidth]{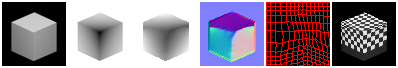}}\\

        \rotatebox{90}{\hspace{0.2cm}$\alpha=0.7$} & &
        \multicolumn{5}{c}{\includegraphics[trim={0 0 2.4cm 0cm},clip, width=0.89\linewidth]{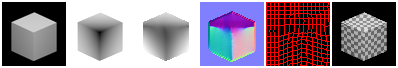}}\\

        \rotatebox{90}{\hspace{0.2cm}w/o MS} & \rotatebox{90}{\hspace{0.05cm}tex. coord.} &
        \multicolumn{5}{c}{\includegraphics[trim={0 0 2.4cm 0cm},clip, width=0.89\linewidth]{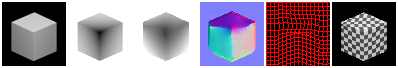}}\\
    \end{tabularx}
    \captionof{figure}{\textbf{Qualitative ablation study for the changes in loss function weights, blend weight, and texture coordinates parametrization.}}
    \label{fig:ablation2}
\end{table}

\section{Additional Ablation Study}
\label{app:ablation}
To evaluate depth reconstruction from the texture coordinates using LSCM, we begin with a cube’s ground-truth depth map and initialize its texture coordinates as an identity grid. We then optimize these coordinates following Equation~\ref{eq:lscm_energy}. Next, we use a spherical depth initialization and attempt to recover the original depth map using the optimized texture coordinates. As illustrated in Figure~\ref{fig:depth-rec-verification}, this approach successfully reconstructs the depth, confirming that depth can be reliably inferred from texture coordinates within the LSCM energy framework.

In Figure~\ref{fig:ablation2} and Table~\ref{table:ablation2}, we present the ablation study on the possible variations in our model design, such as loss function weights, blend weight, and the effect of removing the multi-scale optimization of the texture coordinates.

Furthermore, we show a qualitative ablation study on the impact of input image cues on additional objects not shown in the main paper in Figures~\ref{fig:cuecylinder},~\ref{fig:cuepyramid}, and~\ref{fig:cuesphere}.

\section{Uncurated Samples Reconstruction}
To ensure a fair evaluation and rule out cherry-picking, we assess our approach on uncurated samples. For each object shown in Figure~\ref{fig:text-to-3d} in the main paper, we generate 10 images using Stable Diffusion with the same text prompt. We then apply our method to these images and present the results in Figures~\ref{fig:uncurated1} and~\ref{fig:uncurated-2}. Our evaluations show that our method is stable, without requiring any hyperparameter tuning, and performs best when the object is centered in the image.

\section{RealFusion Analysis}
\label{app:realfusion}
Since RealFusion (RF)~\cite{MelasKyriazi2023RealFusion3R} performs poorly on some of our input images, we analyzed the cause and verified that we were correctly running the code (\url{https://github.com/lukemelas/realfusion}). To this end, we ran the official implementation on the example input images reported on their paper and webpage using the default configuration, except operating at a $64 \times 64$ resolution to do a fair comparison with DreamFusion. While we were unable to replicate their results exactly, the method generally succeeds (see Figure~\ref{fig:realfusionpaperresults}). Since our test cases were simpler in shape and texture, it suggests that RealFusion requires dense visual features from either texture or small-scale geometric details to succeed. This is logical as multi-view constraints require distinct features that can be matched across views. By contrast, our virtual textures work well on untextured objects yet struggle to reconstruct geometric details as the projected texture is relatively coarse---the two approaches are complementary.
   
\begin{table}[t]
    \centering
    \renewcommand{\arraystretch}{0}  %
    \renewcommand{\tabcolsep}{0pt}
    \begin{tabularx}{0.9\linewidth}{c@{\hspace{0.15cm}}c@{\hspace{0.2cm}}Y*{5}{Y}}
        &
        &
        Input &
        GT &
        Recon. & 
        Recon. \\
        &
        &
        Image &
        Depth &
        Depth &
        Normal \\
        \rotatebox{90} {\hspace{0.7cm}S} &
        &
        \multicolumn{4}{c}{\includegraphics[trim={0 0 4.67cm 0}, clip, width=0.8\linewidth]{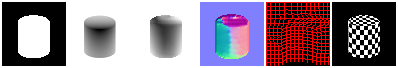}} \\
        \rotatebox{90}{\hspace{0.35cm}S + SH} &
        &
        \multicolumn{4}{c}{\includegraphics[trim={0 0 4.67cm 0}, clip, width=0.8\linewidth]{images/image-to-3d/cylinder.png}} \\
        \rotatebox{90}{\hspace{0.35cm}S + SH} &
        \rotatebox{90}{\hspace{0.35cm}+ RT} &
        \multicolumn{4}{c}{\includegraphics[trim={0 0 4.67cm 0}, clip, width=0.8\linewidth]{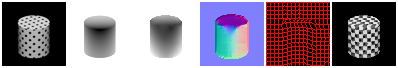}} \\
        \rotatebox{90}{\hspace{0.35cm}S + SH} &
        \rotatebox{90}{\hspace{0.35cm}+ NT} &
        \multicolumn{4}{c}{\includegraphics[trim={0 0 4.67cm 0}, clip, width=0.8\linewidth]{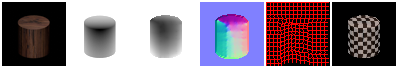}} \\
    \end{tabularx}
    \captionof{figure}{\textbf{Qualitative ablation study on the effect of input image cues for cylinder reconstruction.} We analyze the impact of four different cues in the input image: (S) silhouette, (S + SH) silhouette + shading, (S + SH + RT) silhouette + shading + regular texture, and (S + SH + NT) silhouette + shading + natural texture.}
    \label{fig:cuecylinder}
\end{table}

\begin{table}[t]
    \centering
    \renewcommand{\arraystretch}{0}  %
    \renewcommand{\tabcolsep}{0pt}
    \begin{tabularx}{0.9\linewidth}{c@{\hspace{0.15cm}}c@{\hspace{0.2cm}}Y*{5}{Y}}
        &
        &
        Input &
        GT &
        Recon. & 
        Recon. \\
        &
        &
        Image &
        Depth &
        Depth &
        Normal \\
        \rotatebox{90} {\hspace{0.7cm}S} &
        &
        \multicolumn{4}{c}{\includegraphics[trim={0 0 4.67cm 0}, clip, width=0.8\linewidth]{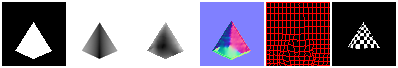}} \\
        \rotatebox{90}{\hspace{0.35cm}S + SH} &
        &
        \multicolumn{4}{c}{\includegraphics[trim={0 0 4.67cm 0}, clip, width=0.8\linewidth]{images/image-to-3d/pyramid.png}} \\
        \rotatebox{90}{\hspace{0.35cm}S + SH} &
        \rotatebox{90}{\hspace{0.35cm}+ RT} &
        \multicolumn{4}{c}{\includegraphics[trim={0 0 4.67cm 0}, clip, width=0.8\linewidth]{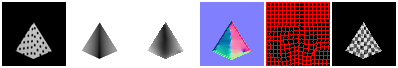}} \\
        \rotatebox{90}{\hspace{0.35cm}S + SH} &
        \rotatebox{90}{\hspace{0.35cm}+ NT} &
        \multicolumn{4}{c}{\includegraphics[trim={0 0 4.67cm 0}, clip, width=0.8\linewidth]{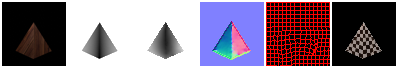}} \\
    \end{tabularx}
    \captionof{figure}{\textbf{Qualitative ablation study on the effect of input image cues for pyramid reconstruction.} We analyze the impact of four different cues in the input image: (S) silhouette, (S + SH) silhouette + shading, (S + SH + RT) silhouette + shading + regular texture, and (S + SH + NT) silhouette + shading + natural texture.}
    \label{fig:cuepyramid}
\end{table}

\begin{table}[t]
    \centering
    \renewcommand{\arraystretch}{0}  %
    \renewcommand{\tabcolsep}{0pt}
    \begin{tabularx}{0.9\linewidth}{c@{\hspace{0.15cm}}c@{\hspace{0.2cm}}Y*{5}{Y}}
        &
        &
        Input &
        GT &
        Recon. & 
        Recon. \\
        &
        &
        Image &
        Depth &
        Depth &
        Normal \\
        \rotatebox{90} {\hspace{0.7cm}S} &
        &
        \multicolumn{4}{c}{\includegraphics[trim={0 0 4.67cm 0}, clip, width=0.8\linewidth]{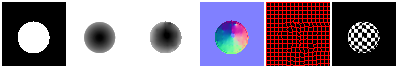}} \\
        \rotatebox{90}{\hspace{0.35cm}S + SH} &
        &
        \multicolumn{4}{c}{\includegraphics[trim={0 0 4.67cm 0}, clip, width=0.8\linewidth]{images/image-to-3d/sphere.png}} \\
        \rotatebox{90}{\hspace{0.35cm}S + SH} &
        \rotatebox{90}{\hspace{0.35cm}+ RT} &
        \multicolumn{4}{c}{\includegraphics[trim={0 0 4.67cm 0}, clip, width=0.8\linewidth]{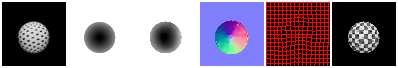}} \\
        \rotatebox{90}{\hspace{0.35cm}S + SH} &
        \rotatebox{90}{\hspace{0.35cm}+ NT} &
        \multicolumn{4}{c}{\includegraphics[trim={0 0 4.67cm 0}, clip, width=0.8\linewidth]{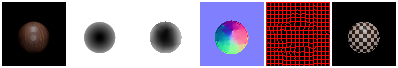}} \\
    \end{tabularx}
    \captionof{figure}{\textbf{Qualitative ablation study on the effect of input image cues for sphere reconstruction.} We analyze the impact of four different cues in the input image: (S) silhouette, (S + SH) silhouette + shading, (S + SH + RT) silhouette + shading + regular texture, and (S + SH + NT) silhouette + shading + natural texture.}
    \label{fig:cuesphere}
\end{table}

\begin{table*}[t!]
    \centering
    \begin{minipage}[t]{0.3\textwidth}
    \renewcommand{\arraystretch}{0}  %
    \renewcommand{\tabcolsep}{0pt}
    \begin{tabularx}{\textwidth}{{Y}*{3}{Y}}
         Input & 
         Recon. & 
         Recon. \\
         &
         Depth &
         Normal
         \vspace{0.2cm}\\
        \multicolumn{3}{c}{\includegraphics[trim={0 0 4.67cm 0cm},clip, width=\linewidth]{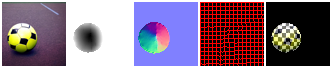}}\\
        \multicolumn{3}{c}{\includegraphics[trim={0 0 4.67cm 0cm},clip, width=\linewidth]{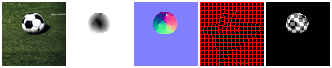}}\\
        \multicolumn{3}{c}{\includegraphics[trim={0 0 4.67cm 0cm},clip, width=\linewidth]{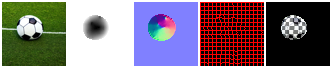}}\\
        \multicolumn{3}{c}{\includegraphics[trim={0 0 4.67cm 0cm},clip, width=\linewidth]{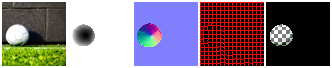}}\\
        \multicolumn{3}{c}{\includegraphics[trim={0 0 4.67cm 0cm},clip, width=\linewidth]{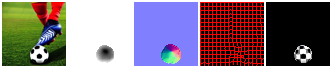}}\\
        \multicolumn{3}{c}{\includegraphics[trim={0 0 4.67cm 0cm},clip, width=\linewidth]{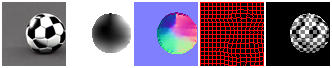}}\\
        \multicolumn{3}{c}{\includegraphics[trim={0 0 4.67cm 0cm},clip, width=\linewidth]{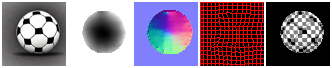}}\\
        \multicolumn{3}{c}{\includegraphics[trim={0 0 4.67cm 0cm},clip, width=\linewidth]{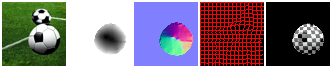}}\\
        \multicolumn{3}{c}{\includegraphics[trim={0 0 4.67cm 0cm},clip, width=\linewidth]{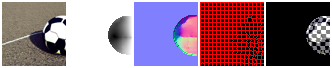}}\\
        \multicolumn{3}{c}{\includegraphics[trim={0 0 4.67cm 0cm},clip, width=\linewidth]{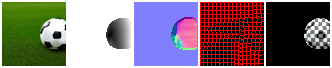}}\\
    \end{tabularx}
    \end{minipage}
    \hfill
    \begin{minipage}[t]{0.3\textwidth}
    \renewcommand{\arraystretch}{0}  %
    \renewcommand{\tabcolsep}{0pt}
    \begin{tabularx}{\textwidth}{{Y}*{3}{Y}}
         Input & 
         Recon. & 
         Recon. \\
         &
         Depth &
         Normal
         \vspace{0.2cm}\\
         \multicolumn{3}{c}{\includegraphics[trim={0 0 4.67cm 0cm},clip, width=\linewidth]{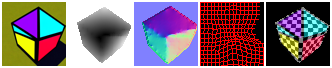}}\\
        \multicolumn{3}{c}{\includegraphics[trim={0 0 4.67cm 0cm},clip, width=\linewidth]{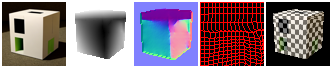}}\\
        \multicolumn{3}{c}{\includegraphics[trim={0 0 4.67cm 0cm},clip, width=\linewidth]{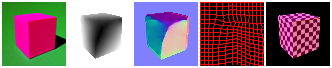}}\\
        \multicolumn{3}{c}{\includegraphics[trim={0 0 4.67cm 0cm},clip, width=\linewidth]{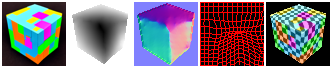}}\\
        \multicolumn{3}{c}{\includegraphics[trim={0 0 4.67cm 0cm},clip, width=\linewidth]{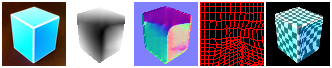}}\\
        \multicolumn{3}{c}{\includegraphics[trim={0 0 4.67cm 0cm},clip, width=\linewidth]{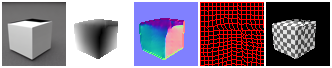}}\\
        \multicolumn{3}{c}{\includegraphics[trim={0 0 4.67cm 0cm},clip, width=\linewidth]{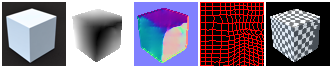}}\\
        \multicolumn{3}{c}{\includegraphics[trim={0 0 4.67cm 0cm},clip, width=\linewidth]{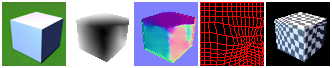}}\\
        \multicolumn{3}{c}{\includegraphics[trim={0 0 4.67cm 0cm},clip, width=\linewidth]{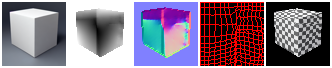}}\\
         \multicolumn{3}{c}{\includegraphics[trim={0 0 4.67cm 0cm},clip, width=\linewidth]{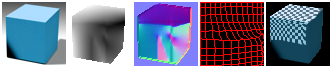}}\\
    \end{tabularx}
    \end{minipage}
    \hfill
    \begin{minipage}[t]{0.3\textwidth}
    \renewcommand{\arraystretch}{0}  %
    \renewcommand{\tabcolsep}{0pt}
    \begin{tabularx}{\textwidth}{{Y}*{3}{Y}}
         Input & 
         Recon. & 
         Recon. \\
         &
         Depth &
         Normal
         \vspace{0.2cm}\\
         \multicolumn{3}{c}{\includegraphics[trim={0 0 4.67cm 0cm},clip, width=\linewidth]{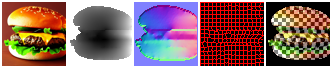}}\\
        \multicolumn{3}{c}{\includegraphics[trim={0 0 4.67cm 0cm},clip, width=\linewidth]{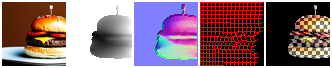}}\\
        \multicolumn{3}{c}{\includegraphics[trim={0 0 4.67cm 0cm},clip, width=\linewidth]{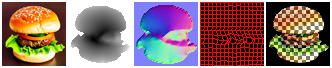}}\\
        \multicolumn{3}{c}{\includegraphics[trim={0 0 4.67cm 0cm},clip, width=\linewidth]{images/uncurated/hamburger/hamburger2.png}}\\
        \multicolumn{3}{c}{\includegraphics[trim={0 0 4.67cm 0cm},clip, width=\linewidth]{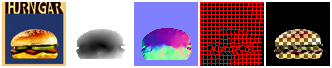}}\\
        \multicolumn{3}{c}{\includegraphics[trim={0 0 4.67cm 0cm},clip, width=\linewidth]{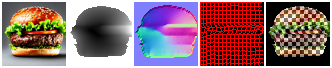}}\\
        \multicolumn{3}{c}{\includegraphics[trim={0 0 4.67cm 0cm},clip, width=\linewidth]{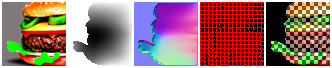}}\\
        \multicolumn{3}{c}{\includegraphics[trim={0 0 4.67cm 0cm},clip, width=\linewidth]{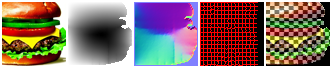}}\\
        \multicolumn{3}{c}{\includegraphics[trim={0 0 4.67cm 0cm},clip, width=\linewidth]{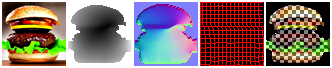}}\\
        \multicolumn{3}{c}{\includegraphics[trim={0 0 4.67cm 0cm},clip, width=\linewidth]{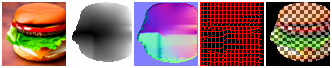}}\\
    \end{tabularx}
    \end{minipage}
    \captionof{figure}{\textbf{Evaluation of our method on 10 uncurated samples of a soccer ball (first column), a cube (second column), and a hamburger (third column).} The input images are generated using Stable Diffusion with the text prompt \emph{a soccer ball}, \emph{a cube}, and \emph{a hamburger}.
}
    \label{fig:uncurated1}
\end{table*}

\begin{table*}[t!]
    \centering
    \begin{minipage}[t]{0.3\textwidth}
    \renewcommand{\arraystretch}{0}  %
    \renewcommand{\tabcolsep}{0pt}
    \begin{tabularx}{\textwidth}{{Y}*{3}{Y}}
         Input & 
         Recon. & 
         Recon. \\
         &
         Depth &
         Normal
         \vspace{0.2cm}\\
        \multicolumn{3}{c}{\includegraphics[trim={0 0 4.67cm 0cm},clip, width=\linewidth]{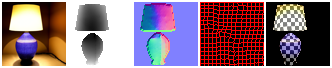}}\\
        \multicolumn{3}{c}{\includegraphics[trim={0 0 4.67cm 0cm},clip, width=\linewidth]{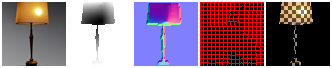}}\\
        \multicolumn{3}{c}{\includegraphics[trim={0 0 4.67cm 0cm},clip, width=\linewidth]{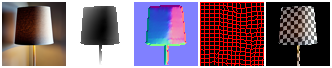}}\\
        \multicolumn{3}{c}{\includegraphics[trim={0 0 4.67cm 0cm},clip, width=\linewidth]{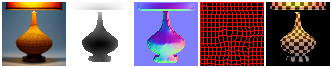}}\\
        \multicolumn{3}{c}{\includegraphics[trim={0 0 4.67cm 0cm},clip, width=\linewidth]{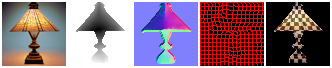}}\\
        \multicolumn{3}{c}{\includegraphics[trim={0 0 4.67cm 0cm},clip, width=\linewidth]{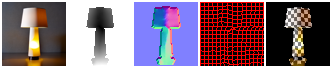}}\\
        \multicolumn{3}{c}{\includegraphics[trim={0 0 4.67cm 0cm},clip, width=\linewidth]{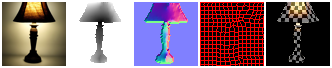}}\\
        \multicolumn{3}{c}{\includegraphics[trim={0 0 4.67cm 0cm},clip, width=\linewidth]{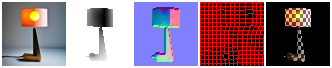}}\\
        \multicolumn{3}{c}{\includegraphics[trim={0 0 4.67cm 0cm},clip, width=\linewidth]{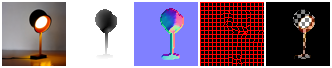}}\\
        \multicolumn{3}{c}{\includegraphics[trim={0 0 4.67cm 0cm},clip, width=\linewidth]{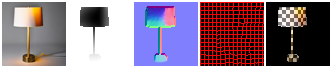}}\\
    \end{tabularx}
    \end{minipage}
    \hfill
    \begin{minipage}[t]{0.3\textwidth}
    \renewcommand{\arraystretch}{0}  %
    \renewcommand{\tabcolsep}{0pt}
    \begin{tabularx}{\textwidth}{{Y}*{3}{Y}}
         Input & 
         Recon. & 
         Recon. \\
         &
         Depth &
         Normal
         \vspace{0.2cm}\\
         \multicolumn{3}{c}{\includegraphics[trim={0 0 4.67cm 0cm},clip, width=\linewidth]{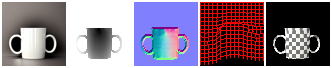}}\\
        \multicolumn{3}{c}{\includegraphics[trim={0 0 4.67cm 0cm},clip, width=\linewidth]{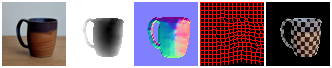}}\\
        \multicolumn{3}{c}{\includegraphics[trim={0 0 4.67cm 0cm},clip, width=\linewidth]{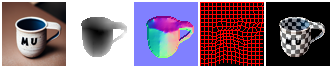}}\\
        \multicolumn{3}{c}{\includegraphics[trim={0 0 4.67cm 0cm},clip, width=\linewidth]{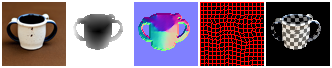}}\\
        \multicolumn{3}{c}{\includegraphics[trim={0 0 4.67cm 0cm},clip, width=\linewidth]{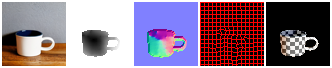}}\\
        \multicolumn{3}{c}{\includegraphics[trim={0 0 4.67cm 0cm},clip, width=\linewidth]{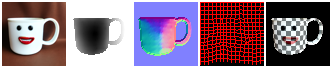}}\\
        \multicolumn{3}{c}{\includegraphics[trim={0 0 4.67cm 0cm},clip, width=\linewidth]{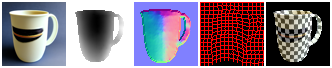}}\\
         \multicolumn{3}{c}{\includegraphics[trim={0 0 4.67cm 0cm},clip, width=\linewidth]{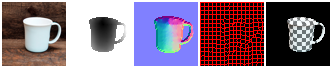}}\\
         \multicolumn{3}{c}{\includegraphics[trim={0 0 4.67cm 0cm},clip, width=\linewidth]{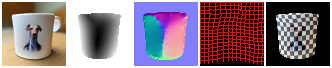}}\\
         \multicolumn{3}{c}{\includegraphics[trim={0 0 4.67cm 0cm},clip, width=\linewidth]{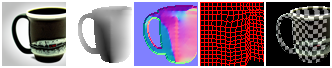}}\\
    \end{tabularx}
    \end{minipage}
    \hfill
    \begin{minipage}[t]{0.3\textwidth}
    \renewcommand{\arraystretch}{0}  %
    \renewcommand{\tabcolsep}{0pt}
    \begin{tabularx}{\textwidth}{{Y}*{3}{Y}}
         Input & 
         Recon. & 
         Recon. \\
         &
         Depth &
         Normal
         \vspace{0.2cm}\\
        \multicolumn{3}{c}{\includegraphics[trim={0 0 4.67cm 0cm},clip, width=\linewidth]{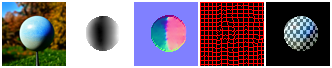}}\\
        \multicolumn{3}{c}{\includegraphics[trim={0 0 4.67cm 0cm},clip, width=\linewidth]{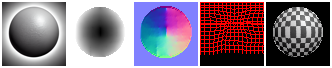}}\\
        \multicolumn{3}{c}{\includegraphics[trim={0 0 4.67cm 0cm},clip, width=\linewidth]{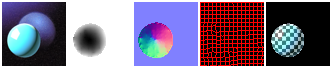}}\\
        \multicolumn{3}{c}{\includegraphics[trim={0 0 4.67cm 0cm},clip, width=\linewidth]{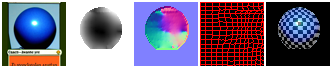}}\\
        \multicolumn{3}{c}{\includegraphics[trim={0 0 4.67cm 0cm},clip, width=\linewidth]{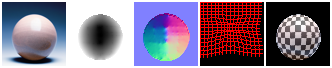}}\\
        \multicolumn{3}{c}{\includegraphics[trim={0 0 4.67cm 0cm},clip, width=\linewidth]{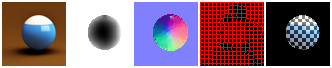}}\\
        \multicolumn{3}{c}{\includegraphics[trim={0 0 4.67cm 0cm},clip, width=\linewidth]{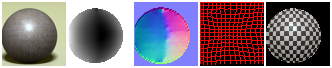}}\\
        \multicolumn{3}{c}{\includegraphics[trim={0 0 4.67cm 0cm},clip, width=\linewidth]{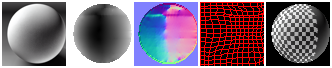}}\\
        \multicolumn{3}{c}{\includegraphics[trim={0 0 4.67cm 0cm},clip, width=\linewidth]{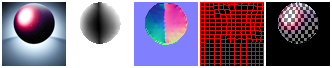}}\\
        \multicolumn{3}{c}{\includegraphics[trim={0 0 4.67cm 0cm},clip, width=\linewidth]{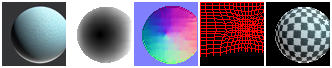}}\\
    \end{tabularx}
    \end{minipage}
    \captionof{figure}{\textbf{Evaluation of our method on 10 uncurated samples of a lamp (first column), a mug (second column), and a sphere (third column).} The input images are generated using Stable Diffusion with the text prompt \emph{a lamp}, \emph{a mug}, and \emph{a sphere}.
}
    \label{fig:uncurated-2}
\end{table*}

\begin{table*}[t!]
    \centering
    \begin{minipage}[t]{0.47\textwidth}
    \renewcommand{\arraystretch}{0}  %
    \renewcommand{\tabcolsep}{0pt}
    \begin{tabularx}{\textwidth}{{Y}*{6}{Y}}
        Input & 
         Recon. & 
         Recon. & 
         Render & 
         Render & 
         Render \\
         & 
         Depth &
         Normal &
         View 1 & 
         View 2 & 
         View 3 \vspace{0.2cm}\\
        \multicolumn{6}{c}{\includegraphics[width=\textwidth]{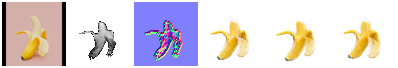}}\\
        \multicolumn{6}{c}{\includegraphics[width=\textwidth]{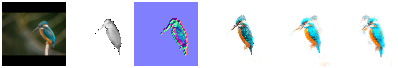}}\\
        \multicolumn{6}{c}{\includegraphics[width=\textwidth]{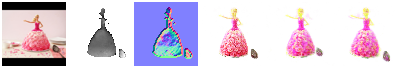}}\\
        \multicolumn{6}{c}{\includegraphics[width=\textwidth]{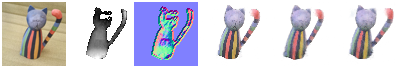}}\\
    \end{tabularx}
    \end{minipage}
    \hfill
    \begin{minipage}[t]{0.47\textwidth}
    \renewcommand{\arraystretch}{0}  %
    \renewcommand{\tabcolsep}{0pt}
    \begin{tabularx}{\textwidth}{{Y}*{6}{Y}}
        Input & 
         Recon. & 
         Recon. & 
         Render & 
         Render & 
         Render \\
         & 
         Depth &
         Normal &
         View 1 & 
         View 2 & 
         View 3 \vspace{0.2cm}\\
        \multicolumn{6}{c}{\includegraphics[width=\textwidth]{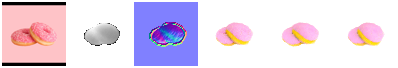}}\\
        \multicolumn{6}{c}{\includegraphics[width=\textwidth]{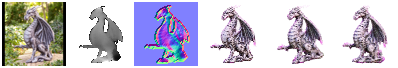}}\\
        \multicolumn{6}{c}{\includegraphics[width=\textwidth]{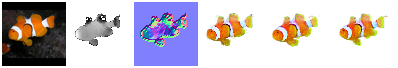}}\\
        \multicolumn{6}{c}{\includegraphics[width=\textwidth]{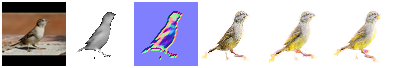}}\\
    \end{tabularx}
    \end{minipage}
    \captionof{figure}{\textbf{Reproduced RealFusion results.} We ran the official implementation of RealFusion on the images provided in the original paper. RealFusion performs well on textured objects. However, its failure on a number of our input images suggests that the multi-view cues alone are insufficient for reconstruction when the object lacks texture and fine-grained geometric features.}
    \label{fig:realfusionpaperresults}
\end{table*}

\end{document}